\documentclass[10pt,twocolumn,letterpaper]{article}

\usepackage{wacv}              

\usepackage{graphicx}
\usepackage{booktabs}
\usepackage{times}
\usepackage{epsfig}
\usepackage{amsmath}
\usepackage{amssymb}
\usepackage{color}
\usepackage{floatrow} 
\usepackage{subcaption}
\usepackage{makecell}
\usepackage{tabularx}
\usepackage{xcolor}

\usepackage{color}

\usepackage{mathrsfs}

\usepackage{enumitem}
\usepackage[pagebackref,breaklinks,colorlinks]{hyperref}

\usepackage[pagebackref,breaklinks,colorlinks]{hyperref}

\usepackage[capitalize]{cleveref}
\crefname{section}{Sec.}{Secs.}
\Crefname{section}{Section}{Sections}
\Crefname{table}{Table}{Tables}
\crefname{table}{Tab.}{Tabs.}

\def\wacvPaperID{*****} 
\def\confName{WACV}
\def\confYear{2025}

\begin{document}

\title{Feature Fusion Transferability Aware Transformer for Unsupervised Domain Adaptation}

\author{Xiaowei Yu\textsuperscript{*}\\
Department of Computer Science and Engineering\\
University of Texas at Arlington\\
{\tt\small xxy1302@mavs.uta.edu}
\and
Zhe Huang\textsuperscript{*}\\
Department of Computer Science\\
Tufts University \\
{\tt\small zh1087@nyu.edu}
\and
Zao Zhang\\
Yiwu Industria \& Commercial College\\
{\tt\small zzchun12826@gmail.com}
}

\maketitle
\renewcommand{\thefootnote}{\relax}
\footnotetext[1]{\textsuperscript{*}Equal contribution.}
\renewcommand{\thefootnote}{\arabic{footnote}}

\begin{abstract}
Unsupervised domain adaptation (UDA) aims to leverage the knowledge learned from labeled source domains to improve performance on the unlabeled target domains. While Convolutional Neural Networks (CNNs) have been dominant in previous UDA methods, recent research has shown promise in applying Vision Transformers (ViTs) to this task. In this study, we propose a novel \textbf{F}eature \textbf{F}usion \textbf{T}ransferability \textbf{A}ware \textbf{T}ransformer (FFTAT) to enhance ViT performance in UDA tasks. Our method introduces two key innovations: First, we introduce a patch discriminator to evaluate the transferability of patches, generating a transferability matrix. We integrate this matrix into self-attention, directing the model to focus on transferable patches. Second, we propose a feature fusion technique to fuse embeddings in the latent space, enabling each embedding to incorporate information from all others, thereby improving generalization. These two components work in synergy to enhance feature representation learning. Extensive experiments on widely used benchmarks demonstrate that our method significantly improves UDA performance, achieving state-of-the-art (SOTA) results.
\end{abstract}

\begin{figure*}[t]
\begin{center}
\includegraphics[scale=0.9]{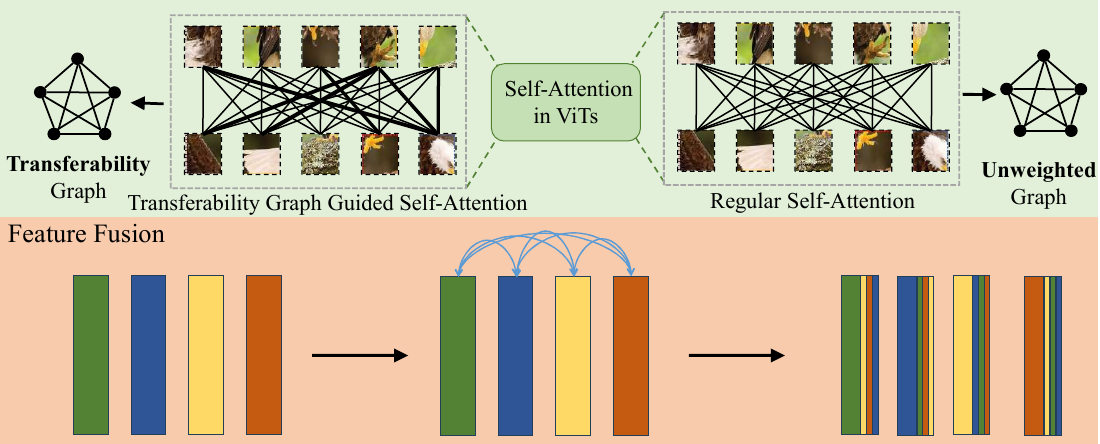}
\end{center}
\caption{ Illustration of transferability graph-guided self-attention and feature fusion. The section above (green) shows the transferability graph guided self-attention and compares it with vanilla self-attention. The section below (orange) illustrates the feature fusion mechanism, where the features of each sample are summed with the features of all other images in the same batch. Each bar represents the features of a sample.}
\label{component_illustration}
\end{figure*}

\section{Introduction}
\label{sec:intro}

Deep neural networks (DNNs) have achieved remarkable breakthroughs across various application fields owing to their impressive automatic feature extraction capabilities. However, such success often relies on the availability of large labeled datasets, which can be challenging to acquire in many real-world scenarios due to the significant time and labor required. Fortunately, unsupervised domain adaptation (UDA) techniques \cite{Wilson20} offer a promising solution by harnessing rich labeled data from a source domain and transferring knowledge to target domains with limited or no labeled examples. The essence of UDA lies in identifying discriminant and domain-invariant features shared between the source domain and target domain within a common latent space~\cite{yang23}. Over the past decade, as interests in domain adaptation research have grown, numerous UDA methods have emerged and evolved~\cite{YU2023UDA, Liang20, Long18}, such as adversarial adaptation, which focuses on discriminating domain-invariant and domain-variant features and acquiring domain-invariant feature representations through adversarial learning~\cite{Zhang19, Long18}. Besides, deep unsupervised domain adaptation techniques usually employ a pre-trained Convolutional Neural Network (CNN) backbone~\cite{LiICCV21}.

Recently, the self-attention mechanism and vision transformer (ViT)~\cite{dosovitskiy2020image, YU2023cpvit, Xiao2024ViT} have received growing interest in the vision community. Unlike convolutional neural networks that gather information from local receptive fields of the given image, ViTs leverage the self-attention mechanism to capture long-range dependencies among patch features through a global view. In ViT and many of its variants, each image is partitioned into a series of non-overlapping fixed-size patches, which are then projected into a latent space as patch tokens and combined with position embeddings. A class token, representing the entire image, is prepended to the patch tokens. All tokens are then fed into a specific number of transformer layers to learn visual representations of the input image. Leveraging the superior global content capture capability of the self-attention mechanism, ViTs have demonstrated impressive performance across various vision tasks, including image classification \cite{dosovitskiy2020image}, video understanding \cite{Girdhar19video}, and object detection \cite{carion2020end}.

Despite increasing interest, only a few studies have explored the application of ViTs for unsupervised domain adaptation tasks \cite{yang23, Sun22, Xu22, YU2023UDA}. In this work, we introduce a novel \textbf{F}eature \textbf{F}usion \textbf{T}ransferability \textbf{A}ware \textbf{T}ransformer, designed for unsupervised domain adaptation. FFTAT builds upon TVT~\cite{yang23}, the first ViT-based UDA model, by introducing two key components: (1) a transferability graph-guided self-attention (TG-SA) mechanism that enhances information from highly transferable features while suppressing information from less transferable features, and (2) a carefully designed features fusion (FF) operation that makes each embedding incorporate information from other embeddings in the same batch. Fig.~\ref{component_illustration} illustrates the transferability graph guided self-attention and feature fusion. 

From a graph view, vanilla self-attention among patches can be seen as an unweighted graph, where the patches are considered as nodes, and the attention between nodes is regarded as the edge connecting them. Unlike vanilla self-attention, our proposed transferability graph guided self-attention is controlled by a weighted graph, where the information communication between highly transferable patches is emphasized via a large-weight edge, and the information communication between less transferable patches is attenuated by a small-weight edge~\cite{YU2023cpvit, Yu2024CP}. The transferability graph is automatically learned and updated through learning iterations in the transferability-aware layer, where we design a patch discriminator to evaluate the transferability of each patch. The TG-SA allows for integrative information processing, facilitating the model to focus on {domain-invariant features shared between domains} and gather important information for domain adaptation. The Feature Fusion (FF) operation enables each embedding to integrate information from other embeddings. Different from recent work PMTrans~\cite{ZhuCPVR23} for unsupervised domain adaptation, our feature fusion occurs in the latent space rather than on the image level.

These two new components synergistically enhance robust feature representation learning and generalization in UDA tasks. Extensive experiments on widely used UDA benchmarks demonstrate that FFTAT significantly improves UDA performance, achieving new state-of-the-art results.
In summary, our contributions are as follows:

\begin{itemize}[leftmargin=*]

\item We introduce a novel transferability graph-guided attention mechanism in ViT architecture for UDA, enhancing performance by promoting attention between highly transferable features while suppressing attention between less transferable ones.

\item We propose a feature fusion technique that enhances feature learning and generalization capabilities for UDA.

\item Our proposed model, FFTAT, integrates transferability graph-guided attention and feature fusion mechanisms, resulting in notable advancements and state-of-the-art performance on widely used UDA benchmarks.

\end{itemize}

\begin{figure*}[t]
\begin{center}
\includegraphics[scale=0.18]{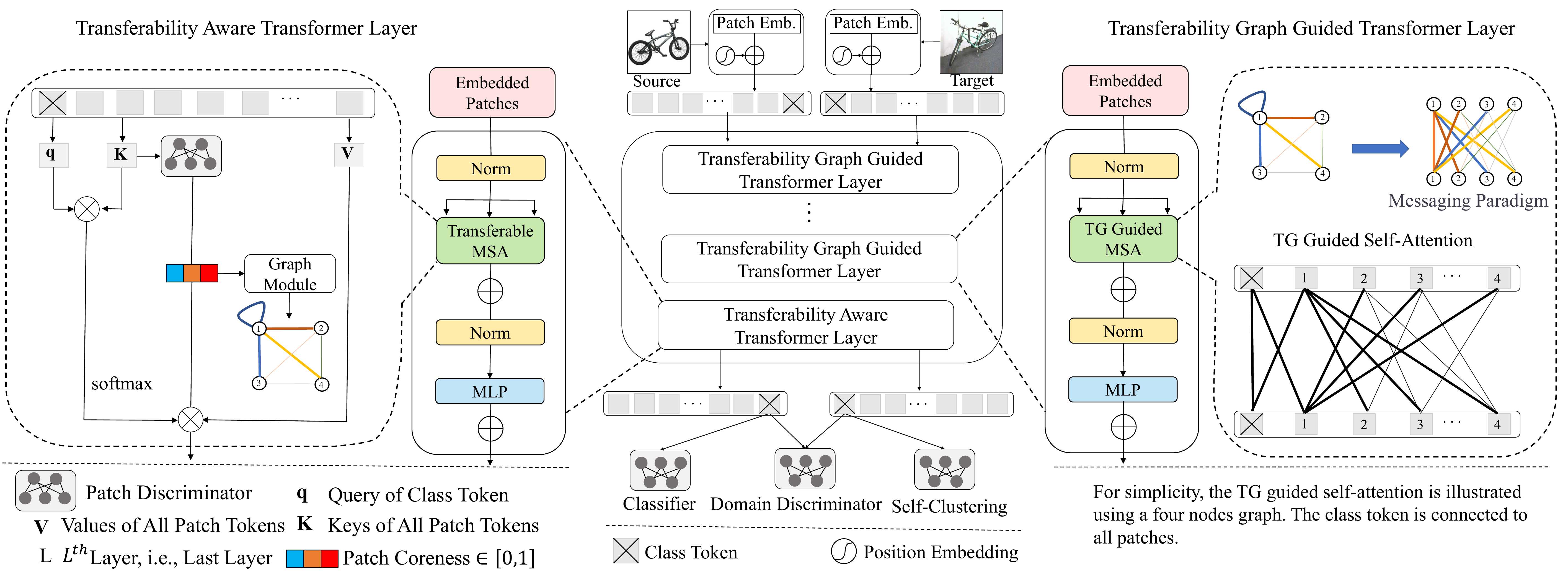}
\end{center}
\caption{The overview of the FFTAT framework. In FFTAT, source and target images are divided into non-overlapping fixed-size patches which are linearly projected into the latent space and concatenated with positional information. A class token is prepended to the image tokens. The tokens are subsequently processed by a transformer encoder. The Feature Fusion Layer mixes the features as illustrated in Fig. \ref{component_illustration}. The patch discriminator assesses the transferability of each patch and generates a transferability graph, which is used to guide the attention mechanism in the transformer layers. The classifier head and self-clustering module operate on source domain images and target domain images, respectively. The Domain Discriminator predicts whether an image belongs to the source or target domain.}
\label{model}
\end{figure*}

\section{Related Work}

\subsection{Unsupervised Domain Adaptation}
UDA aims to learn transferable knowledge across the source and target domains with different distributions \cite{Pan09Survey, Ying18Transfer}.  Various techniques for UDA have been proposed. For example, the discrepancy techniques measure the distribution divergence between source and target domains \cite{Long18, Sun18Coral, Tzeng14Deep}. Adversarial adaptation discriminates domain-invariant and domain-specific representations by playing an adversarial game between the feature extractor and a domain discriminator \cite{Ganin15,yang23}. 
Metric learning aims to optimize new metrics that explicitly capture both the intra-class domain variation and the inter-class domain variation~\cite{Kang2019UDA, Zhu20Deep}. While prevailing UDA methods focus on learning domain-invariant (transferable) features, another line of work emphasizes the importance of domain-specific features~\cite{Sanyal2023Domain}.

\subsection{Data Augmentation}
Data Augmentation has been an integral part of modern deep learning vision models. The general idea is to transform the given data without severely altering the semantics. Common data augmentation techniques include random flip and crop \cite{krizhevsky2012imagenet}, MixUp \cite{zhang2017mixup}, CutOut \cite{devries2017improved}, AutoAugment \cite{hendrycks2019augmix} RandAugment \cite{cubuk2020randaugment} and so on. While these techniques are generally employed in the image space, recent works have explored data augmentation in the embedding space and found promising outcomes in different applications, such as supervised image
classification\cite{verma2019manifold,yu2023exploring} and semi-supervised learning~\cite{Huang2024icml}. Our feature fusion can be viewed as data augmentation in embedding space {for unsupervised domain adaptation}.


\section{Method}
\subsection{Preliminaries}
Let $D_{s} = \left \{ \left ( x_{i}^{s},y_{i}^{s}   \right )  \right \}_{i=1}^{n_{s} }$ represents data in the labeled source domain, where $x_{i}^{s}$ denotes the images, $y_{i}^{s} $ denotes the corresponding labels, and $n_{s}$ denotes the number of samples in labeled source domain. Similarly, let $D_{t} = \left \{ \left ( x_{i}^{t}  \right ) \right \}_{j=1}^{n_{t} }$ represents data in the target domain, consisting $n_{t}$ images but no labels. 
UDA algorithms aim to learn transferable knowledge to minimize domain discrepancy, thereby achieving the desired prediction performance on unlabeled target data. A common approach is to devise an objective function that jointly learns feature embeddings and a classifier. This objective function is formulated as follows:
\begin{equation}\label{GeneralObject}
L = L_{CE}  +\alpha L_{dis}
\end{equation}
where $L_{CE}$ is the standard cross-entropy loss supervised in the source domain, while $L_{dis}$ denotes the divergence loss, which varies in implementation across different algorithms.  Hyperparameter $\alpha$ is used to balance the weight of $L_{dis}$.


\subsection{Overview of FFTAT} \label{FFTATOverview}
We aim to advance ViT-based solutions for Unsupervised Domain Adaptation. Fig.~\ref{model} illustrates the overall framework of our proposed FFTAT. The framework employs a vision transformer backbone, consisting of a domain discriminator for the images (using the class tokens), a patch discriminator for assessing the transferability of the patches, a self-clustering module, and a classifier head. The vision transformer backbone is equipped with our proposed transferability graph-guided attention mechanism (comprising the Transferability Graph Guided Transformer Layer and the Transferability Aware Transformer Layer in Fig. \ref{model}), and a feature fusion mechanism (implemented as the Feature Fusion Layer, as shown in Fig. \ref{model}).

During training, both the source and target images are used. Each image is divided into non-overlapping fixed-size patches which are linearly projected into the latent space and concatenated with positional information. A class token is prepended to the patch tokens. All tokens are subsequently processed by the vision transformer backbone. From a graph view, we consider the patches as nodes and the attention between the patches as edges. This allows us to manipulate the strength of self-attention among patches by adaptively learning a transferability graph during the training process. \textbf{In the first iteration, we initialize the transferability graph as an unweighted graph.} At the end of each iteration, the patch discriminator evaluates the transferability score of the patches and updates the transferability graph. The learned transferability graph is used to rescale the self-attention in the Transferability Aware Transformer Layer and the Transferability Graph Guided Transformer Layers, amplifying information from highly transferable features while damping the less transferable ones.

The Feature Fusion Layer is placed before the Transferability Aware Layer to mix patch token embeddings for better generalization. The classifier head predicts class labels based on the output class token of the labeled source domain images, while the domain discriminator predicts whether images belong to the source or target domain for output class token embeddings of both domains. The self-clustering module utilizes the class token of the target domain to encourage the clustering of learned target domain image representation. Details are introduced below.




\subsection{Transferability Aware Transformer Layer}
The patch tokens correspond to partial regions of the image and capture visual features as fine-grained local representations. Existing work~\cite{yang23, YU2024TMI, Yu2024CPCLIP} shows that the patch tokens are of different semantic importance. In this work, we define the transferability score on a patch to assess its transferability (detailed definition below). A patch with a higher transferability score is more likely to correspond to the highly transferable features. 


To obtain the transferability score of patch tokens, we adopt a patch-level domain discriminator $D_{l} $ to evaluate the local features with a loss function:
\begin{equation}\label{PatchLevelloss}
L_{pat} (x, y^{d})= -\frac{1}{nP} \sum_{x_{i} \in D} \sum_{p=1}^{P} L_{CE} \left ( D_{l} \left ( G_{f} \left ( x_{ip}  \right )   \right ) , y_{ip}^{d}  \right ) 
\end{equation}
where $P$ is the number of patches, $D=D_{s}\cup D_{t}$, $G_{f}$ is the feature encoder, implemented as ViT, $n = n_{s} + n_{t}$, is the total number of images in $D$, $ip$ denotes the $p$ th patch of the $i$ th image, $y_{ip}^{d}$ denotes the domain label of the $p$th token of the $i$th image, i.e., $y_{ip}^{d} = 1$ means source domain, else the target domain. $D_{l}\left ( f_{ip} \right )  $ gives the probability of the patch belonging to the source domain, where $f_{ip}$ denotes the features of the $p$th token of the $i$th image, i.e., $f_{ip} = G_{f} \left ( x_{ip}  \right ) $. During the training process, $D_{l}$ tries to discriminate the patches correctly, assigning 1 to patches from the source domain and 0 to those from the target domain.

Empirically, patches that cannot be easily distinguished by the patch discriminator (e.g., $D_{l}$ is around 0.5) are more likely to correspond to highly transferable features. These highly transferable features capture underlying patterns that remain consistent despite variations in the data distribution between the source and target domains. In the paper, we use the phrase ``transferability'' to capture this property. We define the transferability score of a patch as:


\begin{equation}
\label{transferability_score}
c\left (  f_{ip} \right ) = H\left ( D_{l}\left ( f_{ip}  \right )   \right ) \in \left [ 0, 1 \right ] 
\end{equation}
where $H\left ( \cdot  \right ) $ is the standard entropy function. If the output of the patch discriminator $D_{l}$ is around 0.5, then the transferability score is close to 1, indicating that the features in the patch are highly transferable. A high transferability score means that the features in a patch are highly transferable, and vice versa. Assessing the transferability of patches allows a finer-grained view of the image, separating an image into highly transferable and less transferable patches. Features from highly transferable patches will be amplified while features from less transferable patches will be suppressed.



Let $ \mathcal{C}_{i}= \{ {c}_{i1},..., {c}_{ip} \}$ be the transferability scores of patches of image $i$. The adjacency matrix of the transferability graph can be formulated as:
\begin{equation}\label{CPgraphGeneration}
M_{ts}= \frac{1}{B\mathcal{H}}  \sum_{h=1}^{\mathcal{H}}\sum_{i=1}^{B}      \left [  \mathcal{C}_{i}^{T}  \mathcal{C}_{i} \right ] _{\times }
\end{equation}
$B$ is the batch size, $\mathcal{H}$ is the number of heads, $\left [ \cdot  \right ] _{\times } $ means no gradients back-propagation for the adjacency matrix of the generated transferability graph. 
$M_{ts}$ controls the connection strength of attention between patches.

The vanilla Self-Attention (SA) can then be reformulated as Transferability Aware Self-Attention (TSA) in the Transferability Aware Transformer Layer by integrating with transferability scores:
\begin{equation}\label{SALastLayer}
TSA(q_{cls},K,V) = softmax(\frac{q_{cls}K^{T}}{\sqrt{d} }  )\odot  [ 1;C_{  K_{patch} } ] V
\end{equation}
where $q_{cls}$ is the query of the class token, $K$ represents the key of all tokens, including the class token and patch tokens, $K_{patch}$ is the key of the patch tokens, $C_{ K_{patch}} $ denotes the transferability scores of the patch tokens, $\odot$ is the dot product, and $\left [ ; \right ] $ is the concatenation operation. TSA encourages the class token to take more information from highly transferable patches with higher transferability scores while suppressing information from patches with low transferability scores. The Transferability Aware Multi-head Self-Attention is therefore defined as:
\begin{equation}\label{C-MSA}
T\text{-}MSA(q_{cls},K,V)=Concat(head_{1},...,head_{h} )W^{O}
\end{equation}
where $head_{i}=TSA\left ( q_{cls}W_{i}^{q_{cls}}, KW_{i}^{K} , VW_{i}^{V}   \right ) $. Taking them together, the operations in the transferability aware transformer layer can be formulated as:
\begin{equation}\label{OpeLL}
\begin{split}
& \hat{z}^{l} =T\text{-}MSA\left ( LN\left ( z^{l-1}  \right )  \right ) +z^{l-1}\\
& z^{l}=MLP\left ( LN\left ( \hat{z}^{l} \right )  \right ) + \hat{z} ^{l} 
\end{split}
\end{equation}
where $z^{l-1}$ are output from previous layer. In this way, the Transferability Aware Transformer Layer focuses on fine-grained features that are highly transferable and are discriminative for classification. Here $l=L$, $L$ is the total number of transformer layers in ViT architecture.


\subsection{ Feature Fusion}
\label{sec:LatentFeatureInteraction}
Emerging evidence shows that adding perturbations enhances model robustness \cite{Sun22,Huang2024icml,yu2023exploring}. To enhance the robustness of the generated transferability graphs and to make the model resistant to noisy perturbations, we propose a novel feature fusion technique into our FFTAT framework, implemented as a Feature Fusion Layer, placed before the Transferability Aware Transformer Layer.
Given an image $x_{i}$, let $b_{i}= \{{b_{i1},\cdots,b_{ip}}\}$ denote the embeddings of its patches. 
As illustrated in Fig.~\ref{component_illustration}, each embedding is perturbed by incorporating information from all the other embeddings. We perform the embedding fusion for the source and target domain separately, as instructed in~\cite{yu2023exploring}:
\begin{equation}\label{perturbation}
\left\{\begin{matrix}
 \tilde{b}^{s}_{ip} = \frac{2}{B+1} {b}^{s}_{ip} + \frac{1}{B+1} \sum_{j=1}^{B}{b}^{s}_{jp}  , ~~j\ne i\\
\tilde{b}^{t}_{ip} = \frac{2}{B+1} {b}^{t}_{ip} + \frac{1}{B+1} \sum_{j=1}^{B}{b}^{t}_{jp}  ,~~j\ne i
\end{matrix}\right.
\end{equation}
where $B$ is the batch size, $s$ and $t$ indicate the source and target domain. The FF aids in generating more robust transferability graphs and improves model generalizability. 


\begin{table*}[t]
\caption{ Comparison with SOTA methods on the \textbf{Office-Home} dataset. The best performance is marked in bold. The methods above the horizontal line are CNN-based methods, while the methods below the horizontal line are ViT-based methods.} 
\centering
\resizebox{\linewidth}{!}{
\begin{tabular}{ cccccccccccccc }
\hline
Method        & Ar{\tiny $\rightarrow$}Cl & Ar{\tiny $\rightarrow$}Pr & Ar{\tiny $\rightarrow$}Re & Cl{\tiny $\rightarrow$}Ar & Cl{\tiny $\rightarrow$}Pr & Cl{\tiny $\rightarrow$}Re & Pr{\tiny $\rightarrow$}Ar & Pr{\tiny $\rightarrow$}Cl & Pr{\tiny $\rightarrow$}Re & Re{\tiny $\rightarrow$}Ar & Re{\tiny $\rightarrow$}Cl & Re{\tiny $\rightarrow$}Pr & Avg. \\ \hline
ResNet-50\cite{he2016deep}     & 44.9                & 66.3                & 74.3                & 51.8                & 61.9                & 63.6                & 52.4                & 39.1                & 71.2                & 63.8                & 45.9                & 77.2                & 59.4 \\ 
MinEnt\cite{Grandvalet04}        & 51.0                & 71.9                & 77.1                & 61.2                & 69.1                & 70.1                & 59.3                & 48.7                & 77.0                & 70.4                & 53.0                & 81.0                & 65.8 \\ 
SAFN\cite{XuICCV19}          & 52.0                & 71.7                & 76.3                & 64.2                & 69.9                & 71.9                & 63.7                & 51.4                & 77.1                & 70.9                & 57.1                & 81.5                & 67.3 \\ 
CDAN+E\cite{Long18}        & 54.6                & 74.1                & 78.1                & 63.0                & 72.2                & 74.1                & 61.6                & 52.3                & 79.1                & 72.3                & 57.3                & 82.8                & 68.5 \\ 
DCAN\cite{LiAAAI20}          & 54.5                & 75.7                & 81.2                & 67.4                & 74.0                & 76.3                & 67.4                & 52.7                & 80.6                & 74.1                & 59.1                & 83.5                & 70.5 \\ 
BNM \cite{CuiCVPR20}          & 56.7                & 77.5                & 81.0                & 67.3                & 76.3                & 77.1                & 65.3                & 55.1                & 82.0                & 73.6                & 57.0                & 84.3                & 71.1 \\ 
SHOT\cite{Liang20}          & 57.1                & 78.1                & 81.5                & 68.0                & 78.2                & 78.1                & 67.4                & 54.9                & 82.2                & 73.3                & 58.8                & 84.3                & 71.8 \\ 
ATDOC-NA\cite{LiangCVPR21}      & 58.3                & 78.8                & 82.3                & 69.4                & 78.2                & 78.2                & 67.1                & 56.0                & 82.7                & 72.0                & 58.2                & 85.5                & 72.2 \\ \hline
ViT\cite{dosovitskiy2020image}         & 54.7                & 83.0                & 87.2                & 77.3                & 83.4                & 85.6                & 74.4                & 50.9                & 87.2                & 79.6                & 54.8                & 88.8                & 75.5 \\
TVT\cite{yang23}         & 74.9                & 86.8                & 89.5                & 82.8                & 88.0                & 88.3                & 79.8                & 71.9                & 90.1                & 85.5                & 74.6                & 90.6                & 83.6 \\ 
CDTrans\cite{Xu22}     & 68.8                & 85.0                & 86.9                & 81.5                & 87.1                & 87.3                & 79.6                & 63.3                & 88.2                & 82.0                & 66.0                & 90.6                & 80.5 \\ 
SSRT \cite{Sun22}       & 75.2                & 89.0                & 91.1                & 85.1                & 88.3                & 90.0                & 85.0                & 74.2                & 91.3                & 85.7                & 78.6                & 91.8                & 85.4 \\
PMTrans \cite{ZhuCPVR23}       &81.3              &92.9               & 92.8                & 88.4              & 93.4              & 93.2             & 87.9               & 80.4                & 93.0                & 89.0               &80.9            & 84.8              & 89.0\\
FFTAT (ours) &  {\textbf{83.2}}   &  \textbf{92.9}    & {\textbf{95.2}}      &  {\textbf{91.1}}       &  {\textbf{93.5}}                & {\textbf{95.2}}     & {\textbf{89.7}}    & {\textbf{85.0}}             & {\textbf{94.9}}   &  {\textbf{93.0}}   &  {\textbf{87.5}}                &  {\textbf{95.9}}    & {\textbf{91.4}} \\ \hline
\end{tabular}
}
\label{Office-Home}
\end{table*}

\begin{table*}[t]
\caption{ Comparison with SOTA methods on \textbf{Visda-2017}. The best performance is marked in bold. The methods above the horizontal line are CNN-based methods, while the methods below the horizontal line are ViT-based methods.}
\centering
\scalebox{0.95}{
\begin{tabular}{ cccccccccccccc }
\hline
Method     & plane & bcycl & bus & car & horse & knife & mcycl & person & plant & sktbrd & train & truck & Avg. \\ \hline
ResNet-50\cite{he2016deep}     & 55.1                & 53.3              & 61.9              & 59.1               &80.6                &17.9             &  79.7               & 31.2                &  81.0               & 26.5                &  73.5               & 8.5                & 52.4 \\ 
DANN\cite{Ganin15}          & 81.9              & 77.7                &  82.8                & 44.3                &  81.2                & 29.5                &  65.1                & 28.6                &  51.9              & 54.6                & 82.8                &7.8                & 57.4\\ 
MinEnt\cite{Grandvalet04}        & 80.3               & 75.5                & 75.8               &48.3                &  77.9                & 27.3                & 69.7                & 40.2               & 46.5                & 46.6               & 79.3                &16.0                & 57.0 \\ 
SAFN\cite{XuICCV19}          &93.6                & 61.3                & 84.1               & 70.6               & 94.1               & 79.0               &  91.8                & 79.6               &  89.9                & 55.6                &  89.0              & 24.4              & 76.1 \\ 
CDAN+E\cite{Long18}        & 85.2               &  66.9              & 83.0                &  50.8              & 84.2                & 74.9                & 88.1               &74.5                & 83.4             &76.0              &81.9              &  38.0                & 73.9 \\ 
BNM \cite{CuiCVPR20}          &89.6            & 61.5                &  76.9                & 55.0             & 89.3             & 69.1                & 81.3             & 65.5            &  90.0               & 47.3              & 89.1            & 30.1              &70.4 \\ 
CGDM\cite{DuCVPR21}      & 93.7              &82.7               & 73.2             & 68.4               & 92.9                & 94.5              &88.7                & 82.1             &93.4                & 82.5                & 86.8                & 49.2               & 82.3 \\ 
SHOT\cite{Liang20}          &94.3               & 88.5               & 80.1                 & 57.3              &  93.1             &  93.1              & 80.7               & 80.3              & 91.5               &89.1             &  86.3           & 58.2            & 82.9 \\ 
\hline
ViT~\cite{dosovitskiy2020image}         & 97.7              & 48.1                & 86.6              & 61.6               & 78.1              &  63.4               & 94.7                & 10.3               & 87.7            &   47.7                & 94.4                &  35.5              & 67.1 \\
TVT~\cite{yang23}         &   92.9           & 85.6               &77.5               & 60.5              &  93.6             & 98.2            &89.4                & 76.4               & 93.6               & 92.0                & 91.7                & 55.7               & 83.9 \\ 
CDTrans~\cite{Xu22}     & 97.1                & 90.5               &82.4              & 77.5               & 96.6             & 96.1               &  93.6                &{{88.6}}               &  \textbf{97.9}               & 86.9               &  90.3             & {62.8}             & 88.4 \\ 
SSRT~\cite{Sun22}       & {{98.9}}               & 87.6              & 89.1         &{ \textbf{84.8} }     & 98.3              & 98.7              &96.3              & 81.1               &94.9             & {97.9}           & 94.5                & 43.1               & 88.8 \\
PMTrans \cite{ZhuCPVR23}  & 99.4                & 88.3          & 88.1           & 78.9                & 98.8   &98.3 & 95.8     & 70.3             & 94.6             & 98.3        & 96.3                & 48.5            & 88.0 \\ 
FFTAT (ours) &  {\textbf{99.7}}   &  {\textbf{98.5}}     &{ \textbf{93.1}}    &  81.1     &  {\textbf{99.8}}                & {\textbf{99.5}}     & {\textbf{97.8}}    & {\textbf{89.6}}    & 95.7   &  {\textbf{99.8}}   &  {\textbf{98.7}}                &  {\textbf{72.4}}    & {\textbf{93.8}} \\ \hline
\end{tabular}
}
\label{Visda17}
\end{table*}

\subsection{Transferability Graph Guided Transformer Layer}
As introduced in Section~\ref{FFTATOverview}, we consider the patches as nodes and the attention between patches as edges.
The learned transferability information can be effectively and conveniently integrated into the self-attention mechanism by updating the graph, as illustrated in the right part of Fig. \ref{model}. With the guidance of the transferability graph, the self-attention in Transferability Graph Guided Transformer Layers will focus on the patches with more transferable features, thus steering the model to learn transferable knowledge across the source and target domain.

The transferability graph can be represented by $ \mathcal{G} = ( \mathcal{V}, \mathcal{E})$, with nodes $ \mathcal{V}= \{ {\nu}_{1},..., {\nu}_{n} \}$, edges $ \mathcal{E} = \{ ({\nu}_{p}, {\nu}_{\tilde{p}} )| {\nu}_{p}, {\nu}_{\tilde{p}} \in \mathcal{V} \} $, and adjacency matrix $M_{ts}$. The transferability graph-guided self-attention for a specific patch $p$ at $l$-th layer in the Transferability Graph Guided Transformer Layer is:
\begin{equation}\label{patchCP}
b_{p}^{(l+1)}= \sigma^{(l)}( \frac{q_{p}^{(l)} (K_{\tilde{p}}^{(l)})^{T}  }{\sqrt{d_{k} } })\odot C_{ K_{\tilde{p}}^{(l)} } V_{\tilde{p}}^{(l)} ,~~ \tilde{p} \in N(p) \cup p
\end{equation}
where $\sigma(\cdot)$ is the activation function, which is usually the softmax function in ViTs, $q_{p} ^ {(l)} $ is the query of the $p$ th patch ($p$ th node in $ \mathcal{G} $),  $N(p)$ are the neighborhood nodes of node $p$, $d_k$ is the scale factor with the same dimension of queries and keys, $K_{\tilde{p}}^{(l)}$ and $V_{\tilde{p}}^{(l)}$ are the key and value of nodes $\tilde{p}$, and $C_{K_{\tilde{p}}^{(l)} }$ denotes the transferability scores. Therefore, the transferability graph guided self-attention that is conducted at
the patch level can be formulated as:
\begin{equation}\label{PatchCPSA}
{TG-}SA(Q,K,V,M_{ts})=softmax(\frac{QK^T\odot M_{ts}}{\sqrt{d_k} })V
\end{equation}
where queries, keys, and values of all patches are packed into matrices $Q$, $K$, and $V$, respectively, $M_{ts}$ is the adjacency matrix defined in Eq~\ref{CPgraphGeneration}. The transferability graph guided multi-head attention is then formulated as:
\begin{equation}
\begin{aligned}
MSA(Q,K,V,M_{ts})=Concat(head_{1}, ...,head_{h})W^{o}\\
\end{aligned}
\end{equation}
where $ head_{i}=TG-SA(  QW_{i}^{Q}, KW_{i}^{K} , VW_{i}^{V},M_{ts} ) $
. The learnable parameter matrices $W_{i}^{Q}$, $W_{i}^{K}$, $W_{i}^{V}$ and $W^{O}$ are the projections. Multi-head attention helps the model to jointly aggregate information from different representation subspaces at various positions. In this work, we apply the transferability guidance to each representation subspace.

\subsection{Overall Objective Function }
Since our proposed FFTAT has a classifier head, a self-clustering module, a patch discriminator, and a domain discriminator, there are four terms in the overall objective function. The classification loss is formulated as:
\begin{equation}\label{clcloss}
L_{clc} \left ( x^{s}, y^{s}   \right ) = \frac{1}{n_{s}}\sum_{x_{i}\in D_{s}  }  L_{CE} \left ( G_{c} \left ( G_{f} \left ( x_{i}^{s} \right )  \right ) , y_{i}^{s} \right ) 
\end{equation}
where $G_{c}$ is the classifier head, and $G_{f}$ is the feature extractor, i.e., the ViT with transferability graph-guided self-attention and feature fusion in our work.

The domain discriminator takes the class token of images from the source and target domain and tries to discriminate the class token, i.e., the representation of the entire image, to the source or target domain:
\begin{equation}\label{domainloss}
L_{dis} (x, y^{d})  = -\frac{1}{n} \sum_{x_{i} \in D}L_{ce}\left ( D_{g}\left ( G_{f}\left ( x_{i} \right ) , y_{i}^{d} \right )  \right )  
\end{equation}
where $D{g}$ is the domain discriminator, and $y_{i}^{d}$ is the the domain label (i.e., $y_{i}^{d}= 1$ means source domain, $y_{i}^{d}= 0$ is target).

The self-clustering module is inspired by the cluster assumption \cite{Chapelle05} for images from the target domain without labels and the probability $\mathrm{p}^{t}=softmax\left ( G_{c}\left ( G_{f}\left ( x^{t} \right )  \right )  \right ) $ of target image $x^{t}$ is optimized to maximize the mutual information with $x^{t}$ \cite{yang23}. The self-clustering loss term is formulated as:
\begin{equation}\label{self-cluseringloss}
I\left ( \mathrm{p}^{t}; x^{t} \right ) = H\left ( \bar{\mathrm{p}^{t}}  \right ) -\frac{1}{n_{t}} \sum_{i=1}^{n_{t}}H\left ( \mathrm{p}_{i}^{t} \right )  
\end{equation}
where $\bar{\mathrm{p}^{t}}=\mathbb{E}\left [ \mathrm{p}^{t} \right ] $. The self-clustering loss encourages the model to learn clustered target features.

\begin{table*}[!htp] \tiny
\caption{ Comparison with SOTA methods on \textbf{DomainNet}. The best performance is marked in bold. \textbf{PMTrans takes advantage of a different data partitioning strategy. Our FFTAT adheres to the same data partitioning strategy as SSRT and CDTrans.} } 

\resizebox{\linewidth}{!}{
\begin{subfloatrow}
\setlength{\tabcolsep}{0.6mm}{} 
\begin{tabular}[t]{cccccccc}
  \hline
 \makecell[c]{ResNet101 \\ \cite{he2016deep}} & clp &  inf & pnt & qdr & rel &  skt & Avg.
  \\
  \hline
 clp & - & 19.3 & 37.5 & 11.1 & 52.2 & 41.1 & 32.2   \\
  \hline
 inf & 30.2 & - & 31.2 & 3.6 & 44.0 & 27.9 & 27.4   \\
  \hline
 pnt & 39.6 & 18.7 & - & 4.9 & 54.5 & 36.3 & 30.8  \\
  \hline
 qdr & 7.0 & 0.9 & 1.4 & - & 4.1 & 8.3 & 4.3  \\
  \hline
 rel & 48.4 & 22.2 & 49.4 & 6.4 & - & 38.8 & 33.0  \\
   \hline
 skt & 46.9 & 15.4 & 37.0 & 10.9 & 47.0 & - &  31.4  \\
  \hline
Avg. & 34.4 & 15.3 & 31.3 & 7.4 & 40.4 & 30.5 & 26.6  \\
  \hline
\end{tabular}

\begin{tabular}[t]{cccccccc}
  \hline
 \makecell[c]{MIMTFL\\\cite{GaoECCV20}} & clp &  inf & pnt & qdr & rel &  skt & Avg.
  \\
  \hline
  clp & - & 15.1 & 35.6 & 10.7 & 51.5& 43.1 & 31.2   \\
  \hline
 inf & 32.1 & - & 31.0 & 2.9 & 48.5 & 31.0 & 29.1   \\
  \hline
 pnt & 40.1 & 14.7 & - & 4.2 & 55.4 & 36.8 & 30.2  \\
  \hline
 qdr & 18.8 & 3.1 & 5.0 & - & 16.0 & 13.8 & 11.3  \\
  \hline
 rel & 48.5 & 19.0 & 47.6 & 5.8 & - & 39.4 & 22.1  \\
   \hline
 skt & 51.7 & 16.5 & 40.3 & 12.3 & 53.5 & - &  34.9  \\
  \hline
Avg. & 38.2 & 13.7 & 31.9 & 7.2 & 45.0 & 32.8 & 28.1  \\
  \hline
\end{tabular}

\begin{tabular}[t]{cccccccc}
  \hline
  \makecell[c]{CGDM\\ \cite{DuCVPR21} }& clp &  inf & pnt & qdr & rel &  skt & Avg.
  \\
  \hline
  clp & - & 16.9 & 35.3 & 10.8 & 53.5 & 36.9 & 30.7   \\
  \hline
 inf & 27.8 & - & 28.2 & 4.4 & 48.2 & 22.5 & 26.2   \\
  \hline
 pnt & 37.7 & 14.5 & - & 4.6 & 59.4 & 33.5 & 30.0  \\
  \hline
 qdr & 14.9 & 1.5 & 6.2 & - & 10.9 & 10.2 &8.7  \\
  \hline
 rel & 49.4 & 20.8 & 47.2 & 4.8 & - & 38.2 & 32.0  \\
   \hline
 skt & 50.1 & 16.5 & 43.7 & 11.1 & 55.6 & - &  35.4  \\
  \hline
Avg. & 36.0 & 14.0 & 32.1 & 7.1 & 45.5 & 28.3 & 27.2  \\
  \hline
\end{tabular}
\end{subfloatrow}
}

\resizebox{\linewidth}{!}{
\begin{subfloatrow}
\setlength{\tabcolsep}{0.6mm}{} 
\begin{tabular}[t]{cccccccc}
  \hline
 \makecell[c]{MDD+SCDA\\ \cite{LiICCV21} } & clp &  inf & pnt & qdr & rel &  skt & Avg.
  \\
  \hline
  clp & - & 20.4 & 43.3 & 15.2 & 59.3 & 46.5 & 36.9   \\
  \hline
 inf & 32.7 & - & 34.5 & 6.3 & 47.6 & 29.2 & 30.1  \\
  \hline
 pnt & 46.4 & 19.9 & - & 8.1 & 58.8 & 42.9 & 35.2  \\
  \hline
 qdr & 31.1 & 6.6 & 18.0 & = & 28.8 & 22.0 & 21.3  \\
  \hline
 rel & 55.5 & 23.7 & 52.9 & 9.5 & - & 45.2 & 37.4  \\
   \hline
 skt & 55.8 & 20.1 & 46.5 & 15.0 & 56.7 & - &  38.8  \\
  \hline
Avg. & 44.3 & 18.1 & 39.0 & 10.8 & 50.2 & 37.2 & 33.3  \\
  \hline
\end{tabular}

\begin{tabular}[t]{cccccccc}
  \hline
 \makecell[c]{ViT-Base \\ \cite{dosovitskiy2020image} } & clp &  inf & pnt & qdr & rel &  skt & Avg.
  \\
  \hline
  clp & - & 27.2 & 53.1 & 13.2 & 71.2 & 53.3 & 43.6   \\
  \hline
 inf & 51.4 & - & 49.3 & 4.0 & 66.3 & 41.1 & 42.4 \\
  \hline
 pnt & 53.1 & 25.6 & - & 4.8 & 70.0 & 41.8 & 39.1  \\
  \hline
 qdr & 30.5 & 4.5 & 16.0 & - & 27.0 & 19.3 & 19.5  \\
  \hline
 rel & 58.4 & 29.0 & 60.0 & 6.0 & - & 45.8 & 39.9  \\
   \hline
 skt & 63.9 & 23.8 & 52.3 & 14.4 & 67.4 & - &  44.4  \\
  \hline
Avg. & 51.5 & 22.0 & 46.1 & 8.5 & 60.4 & 40.3 & 38.1  \\
  \hline
\end{tabular}

\begin{tabular}[t]{cccccccc}
  \hline
 \makecell[c]{CDTrans\\ \cite{Xu22} } & clp &  inf & pnt & qdr & rel &  skt & Avg.
  \\
  \hline
  clp & - & 29.4 & 57.2 & 26.0 & 72.6 & 58.1 & 48.7   \\
  \hline
 inf & 57.0 & - & 54.4 & 12.8 & 69.5 & 48.4 & 48.4   \\
  \hline
 pnt & 62.9 & 27.4 & - & 15.8 & 72.1 & 53.9 & 46.4  \\
  \hline
 qdr & 44.6 & 8.9 & 29.0 & - & 42.6 & 28.5 & 30.7  \\
  \hline
 rel & 66.2 & 31.0 & 61.5 & 16.2 & - & 52.9 & 45.6  \\
   \hline
 skt & 69.0 & 29.6 & 59.0 & 27.2 & 72.5 & - &  51.5  \\
  \hline
Avg. & 59.9 & 25.3 & 52.2 & 19.6 & 65.9 & 48.4 & 45.2  \\
  \hline
\end{tabular}
\end{subfloatrow}
}

\resizebox{\linewidth}{!}{
\begin{subfloatrow}
\setlength{\tabcolsep}{0.85mm}{} 
\begin{tabular}[t]{cccccccc}
  \hline
  \makecell[c]{SSRT\\ \cite{Xu22} } & clp &  inf & pnt & qdr & rel &  skt & Avg.
  \\
  \hline
  clp & - & 33.8 & 60.2 & 19.4 & 75.8 & 59.8 & 49.8   \\
  \hline
 inf & 55.5 & - & 54.0 & 9.0 & 68.2 & 44.7 & 46.3   \\
  \hline
 pnt & 61.7 & 28.5 & - & 8.4 & 71.4 & 55.2 & 45.0  \\
  \hline
 qdr & 42.5 & 8.8 & 24.2 & - & 37.6 & 33.6 & 29.3  \\
  \hline
 rel & 69.9 & 37.1 & 66.0 & 10.1 & - & 58.9 & 48.4  \\
   \hline
 skt & 70.6 & 32.8 & 62.2 & 21.7 & 73.2 & - &  52.1  \\
  \hline
Avg. & 60.0 & 28.2 & 53.3 & 13.7 & 65.3 & 50.4 & 45.2  \\
  \hline
\end{tabular}

\begin{tabular}[t]{cccccccc}
  \hline
 \makecell[c]{PMTrans* \\ \cite{ZhuCPVR23} } & clp &  inf & pnt & qdr & rel &  skt & Avg.
  \\
  \hline
  clp & - & 34.2 & 62.7 & 32.5 & 79.3 & 63.7 & 54.5   \\
  \hline
 inf &  67.4 & - & 61.1 & 22.2& 78.0 & 57.6 & 57.3   \\
  \hline
 pnt & 69.7 & 33.5 & - & 23.9 & 79.8 & 61.2 & 53.6  \\
  \hline
 qdr & 54.6 & 17.4 & 38.9 & - & 49.5 & 41.0 & 40.3  \\
  \hline
 rel & 74.1 & 35.3 &  70.0 & 25.4 & - & 61.1 & 53.2  \\
   \hline
 skt & 73.8 & 33.0 & 62.6 & 30.9 & 77.5 & - &  55.6 \\
  \hline
Avg. & 67.9 & 30.7 & 59.1 & 27.0 & 72.8 & 56.9 & \textbf{52.4}  \\
  \hline
\end{tabular}

\begin{tabular}[t]{cccccccc}
  \hline
 \makecell[c]{FFTAT \\(ours) } & clp &  inf & pnt & qdr & rel &  skt & Avg.
  \\
  \hline
  clp & - & 39.4 & 70.3 & 25.5 & 81.9 & 70.9 & 57.6   \\
  \hline
 inf & 67.4 & - & 65.9 & {12.6} & 79.4 & 60.0 & 57.1  \\
  \hline
 pnt & 71.9 & 37.9 & - & {11.4} & 81.5 & 65.6 & 53.6  \\
  \hline
 qdr & 43.5 & 12.4 & 28.9 & - & {41.5} & 32.3 & 31.7  \\
  \hline
 rel & 77.7 & 37.2 & 74.3 & { 14.2} & - & 64.2 & 53.5  \\
   \hline
 skt & 78.4 & 34.9 & 70.1 & 28.4 & 78.1 & - &  58.0  \\
  \hline
Avg. & 67.8 & 32.4 & 61.9 & 18.4 & 72.5 & 58.6 & {\emph{51.9}}  \\
  \hline
\end{tabular}
\end{subfloatrow}
}
\label{DomainNet}
\end{table*}

\begin{table}[t] \small
\caption{ Comparison with SOTA methods on \textbf{Office-31}. The best performance is marked in bold. The methods above the horizontal line are CNN-based methods, while the methods below the horizontal line are ViT-based methods.} 
\centering
\setlength{\tabcolsep}{0.1mm}{} 
\begin{tabular}{cccccccc}
\hline
Method     & A{$\rightarrow$}W  & D{$\rightarrow$}W  & W{$\rightarrow$}D   & A{$\rightarrow$}D  & D{$\rightarrow$}A  & W{$\rightarrow$}A  & Avg. \\ 
ResNet-50~\cite{he2016deep}   & 68.4 & 96.7 & 99.3  & 68.9 & 62.5 & 60.7 & 76.1 \\ 
DANN~\cite{Ganin15}        & 82.0 & 96.9 & 99.1  & 79.7 & 68.2 & 67.4 & 82.2 \\ 
rRGrad+CAT~\cite{DengICCV19} & 94.4 & 98.0 & 100.0 & 90.8 & 72.2 & 70.2 & 87.6 \\ 
SAFN+ENT\cite{XuICCV19}   & 90.1 & 98.6 & 99.8  & 90.7 & 73.0 & 70.2 & 87.1 \\ 
CDAN+TN~\cite{WangNIPS19}   & 95.7 & 98.7 & 100.0 & 94.0 & 73.4 & 74.2 & 89.3 \\ 
TAT~\cite{LiuICML19}       & 92.5 & 99.3 & 100.0 & 93.2 & 73.1 & 72.1 & 88.4 \\ 
SHOT \cite{Liang20}      & 90.1 & 98.4 & 99.9  & 94.0 & 74.7 & 74.3 & 88.6 \\ 
MDD+SCDA~\cite{LiICCV21}   & 95.3 & 99.0 & 100.0 & 95.4 & 77.2 & 75.9 & 90.5 \\ \hline
ViT~\cite{dosovitskiy2020image}      & 91.2 & 99.2 & 100.0 & 93.6 & 80.7 & 80.7 & 91.1 \\ 
TVT~\cite{yang23}     & 96.4 & 99.4 & 100.0 & 96.4 & 84.9 & 86.1 & 93.9 \\ 
CDTrans~\cite{Xu22}  & 96.7 & 99.0 & 100.0 & 97.0 & 81.1 & 81.9 & 92.6 \\ 
SSRT~\cite{Sun22}    & 97.7 & 99.2 & 100.0 & {98.6} & 83.5 & 82.2 & 93.5 \\ 
PMTrans~\cite{ZhuCPVR23}        & \textbf{99.5} & {\textbf{99.4}} & 100.0 & \textbf{99.8} & 86.7 & 86.5 & 95.3 \\ 
FFTAT (ours)       & {97.6} & {{99.2}} & 100.0 & 97.8 & {\textbf{90.2}} & {\textbf{91.0}} & {\textbf{96.0}} \\ \hline
\end{tabular}
\label{office31}
\end{table}

Take classification loss (Eq. \ref{clcloss}), domain discrimination loss (Eq. \ref{domainloss}), patch discrimination loss (Eq. \ref{PatchLevelloss}), and self-clustering loss (Eq. \ref{self-cluseringloss}) together, the overall objective function is therefore formulated as:
\begin{equation}
L_{clc} \left ( x^{s}, y^{s}   \right ) + \alpha L_{dis}\left ( x, y^{d}  \right ) + \beta L_{pat} (x, y^{d}) - \gamma I\left ( p^{t}; x^{t} \right )
\end{equation}
where $\alpha$, $\beta$, and $\gamma$ are the hyperparameters that control the influence of each term on the overall function.

\section{Experiments}
\subsection{Dataset and Setting}
We evaluate our proposed FFTAT on widely used UDA benchmarks, including \textbf{Office-31}~\cite{Saenko10}, \textbf{Office-Home}~\cite{Venkateswara17}, \textbf{Visda-2017}~\cite{Peng17}, and \textbf{DomainNet}~\cite{Peng19}. \textbf{Office-31} contains 4,652 images of 31 categories collected from three domains, i.e., Amazon (A), DSLR (D), and Webcam (W). \textbf{Office-Home} has 15,500 images of 65 classes from four domains: Artistic (Ar), Clip Art (Cl), Product (Pr), and Real-world (Re) images. \textbf{Visda2017} is a simulation-to-real dataset, with more than 0.2 million images in 12 classes. \textbf{DomainNet} dataset has the largest scale containing around 0.6 million images of 345 classes in 6 domains: Quickdraw (qdr), Real (rel), Sketch (skt), Clipart (clp), Infograph (inf), Painting (pnt). 

We use the ViT-base with a 16×16 patch size  (ViT-B/16) \cite{dosovitskiy2020image} \cite{Steiner21}, pre-trained on ImageNet~\cite{Russakovsky15}, as the backbone. We use minibatch Stochastic Gradient Descent (SGD) optimizer~\cite{Ruder18} with a momentum of 0.9 as the optimizer. The batch size is set to 16 for {Office-31}, {Office-Home}, {Visda-2017}, and 
{DomainNet} by default. We initialized the learning rate as 0 and linearly warm up to 0.06 after 500 training steps then a cosine decay strategy is applied. For small to middle-scale datasets Office-31 and Office-Home, the training step is set to 5000. For large-scale datasets Visda-2017 and DomainNet, the training step is set to 20000. The hyperparameters $\alpha$, $\beta$, and $\gamma$ are set to $[1.0, 0.01, 0.1]$ for Office-31 and Office-Home, and to $[0.1, 0.1, 0.1]$ for Visda-2017 and DomainNet by default, but can be adjusted for optimal performance on specific tasks. 

\begin{figure}[t]  
  \centering
  \includegraphics[scale=0.11]{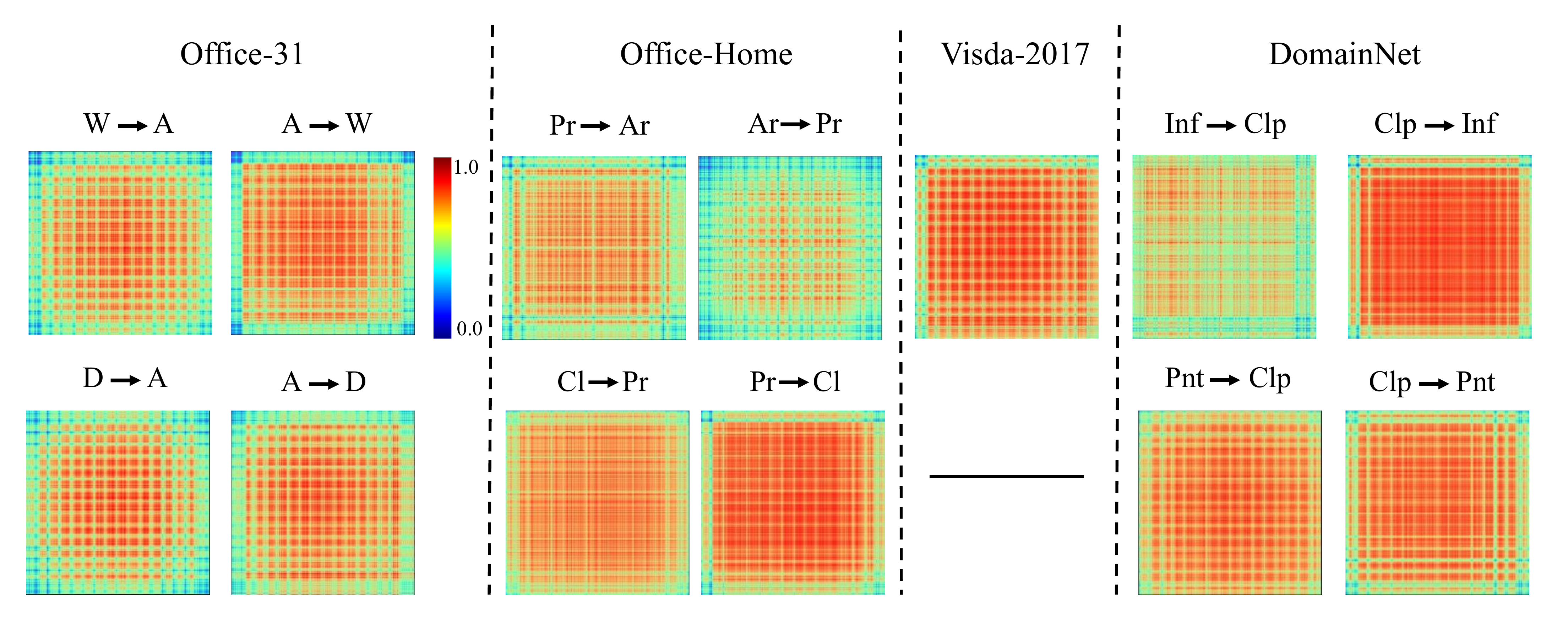}
  \caption{The learned transferability graphs (adjacency matrices) from randomly selected domain adaptation tasks. The weight increases with the intensity of red colors while decreasing with the intensity of blue colors.}
  \label{TGs}
\end{figure}

\subsection{Results}
Table \ref{Office-Home} presents the evaluation results on the Office-Home dataset. The methods listed above the horizontal line are based on CNN architectures, while those below the horizontal line utilize Transformer architectures. 
Compared to its predecessor, TVT~\cite{yang23}, which was the first to adopt the Transformer architecture for UDA, our proposed method, FFTAT, achieves a significant performance improvement of {7.8\%}. FFTAT represents \textbf{the first work to surpass an average accuracy exceeding 90\% on the Office-Home dataset.} Additionally, it is noteworthy that FFTAT outperforms existing methods in each domain adaptation task as well as in the average results by a considerable margin.


Table \ref{Visda17} shows results on the Visda-2017 dataset. We can observe that the FFTAT achieves impressive performance on average results and in most adaptation tasks. Compared to TVT, FFTAT increased performance by 9.9\%. Even when compared to the latest work, PMTrans~\cite{ZhuCPVR23}, our method outperforms it in every task, and by 5.8\% in average accuracy. It is worth noting that PMTrans utilizes the more powerful architecture of SwinTransformer~\cite{liu2021swin} with pre-trained weights from ImageNet-21K. FFTAT is \textbf{the first work to achieve an average accuracy exceeding 93\% on the Visda-2017 dataset, where the performances of other methods are below 89\%.} FFTAT also achieves SOTA results on Office-31, with an average accuracy of 96\% as shown in Table \ref{office31}. 

Experimental results on DomainNet, the largest dataset, are presented in Table \ref{DomainNet}. The performance of the vanilla ViT baseline, standing at 38.1\%, is far from satisfactory. In contrast, FFTAT achieves significant improvements across all domain adaptation tasks, with an average accuracy of 51.9\%. FFTAT surpasses recent Transformer-based methods like CDTrans and SSRT, and remains competitive with the latest work, PMTrans \footnote{At the time of this writing, PMTrans has updated their DomainNet results on their official site to 52.4\% and used a distinct data partitioning strategy that incorporates test data from both source and target domains during training. Our FFTAT adheres to the same data partitioning strategy as SSRT and CDTrans.}.


Overall, the results highlight the superior performance of Transformer-based over CNN-based models, owing to their robust transferable feature representations. Our FFTAT achieves state-of-the-art results across these benchmarks, demonstrating its effectiveness for UDA tasks.

\subsection{Learned Transferability Graphs}
The Transferability Aware Transformer Layer dynamically learns transferability graphs in a data driven manner. To gain insights into the identified patterns, we visualize the adjacency matrices of the learned transferability graphs from two randomly selected domain adaptation tasks across Office-31, Office-Home, and DomainNet datasets. Since the Visda-2017 dataset only consists of two domains (simulation-to-real), there is a single learned transferability graph for Visda-2017. The selected transferability graphs are shown in Fig.~\ref{TGs}. We observe variations in learned transferability patterns for different domain adaptation tasks. For instance, in the Pr to Cl domain adaptation task within the Office-Home dataset, the red-colored area appears large and dense, indicating a significant number of highly transferable patches. Conversely, in the Ar to Pr domain adaptation task, the red-colored area appears sparse and less intense compared to the Pr to Cl task, suggesting fewer highly transferable patches. 

\begin{table}[t] 
\caption{Ablation study on the influence of feature fusion and transferability graph-guided self-attention on model performance across four datasets.} 
\setlength{\tabcolsep}{0.2mm}{} 
\begin{tabular}{c|c|c|c|c}
\hline
                       & Office-31 & Office-Home & Visda2017 & DomainNet \\ \hline
FFTAT                 & 96.0    & 91.4      & 93.8     & 51.9     \\
w/o FF   & 92.7    & 84.5      &84.5     & 43.6     \\
w/o TG-SA & 95.8     & 90.6      & 93.5    & 49.5     \\ \hline
\end{tabular} \label{abalation}
\end{table}

\subsection{Ablation Studies}
We conducted comprehensive ablation studies to assess the impact of the two key components, TG-SA and FF, on model performance across these four datasets. Each component was removed individually to isolate its effect. It is noteworthy that removing the transferability graph-guiding mechanism restores vanilla self-attention in the regular ViT. Table~\ref{abalation} reports the average accuracy across all tasks in each dataset. On average, the removal of Feature fusion or Transferability Graph Guidance leads to performance degradation. These findings demonstrate that each component contributes to the final performance of FFTAT. 

\section{Conclusion}
In this study, we introduce FFTAT, a novel ViT-based solution tailored for unsupervised domain adaptation. FFTAT leverages transferability graphs to guide self-attention (TG-SA) within the Vision Transformer framework, enhancing the emphasis on highly transferable features. and the Feature Fusion (FF) operation to intentionally perturb embeddings, thereby promoting robust feature learning. Extensive experiments demonstrate the efficacy of the Transferability Graph Guided Self-Attention and the Feature Fusion, paving the way for further advancements in this field.

{\small
\bibliographystyle{ieee_fullname}
\bibliography{egbib}
}

\end{document}